# EFFECTIVITY OF SUPER RESOLUTION CONVOLUTIONAL NEURAL NETWORK FOR THE ENHANCEMENT OF LAND COVER CLASSIFICATION FROM MEDIUM RESOLUTION SATELLITE IMAGES

Pritom Bose[1,*], Debolina Halder[2], Oliur Rahman[3], Turash Haque Pial[4,**]

[1,2,3,4,5]Bangladesh University of Engineering and Technology

Email: *bose.me.buet@gmail.com, **pial.buet@gmail.com

**ABSTRACT:** In the modern world, satellite images play a key role in forest management and degradation monitoring. For a precise quantification of forest land cover changes, availability of spatially fine resolution data is a necessity. Since 1972, NASA's LANDSAT Satellites are providing terrestrial images covering every corner of the earth, which have been proved to be a highly useful resource for terrestrial change analysis and has been used in numerous other sectors. However, freely accessible satellite images are, generally, of medium to low resolution which is a major hindrance for the precision of the analysis. Hence, we performed a comprehensive study to prove our point that, enhancement of resolution by Super Resolution Convolutional Neural Network (SRCNN) will lessen the chance of misclassification of pixels, even under the established recognition methods. We tested the method on original LANDSAT-7 images of different regions of Sundarbans and their upscaled versions which were produced by bilinear interpolation, bicubic interpolation and SRCNN respectively and it was discovered that SRCNN outperforms the others by a significant amount.

**INTRODUCTION:**

Multiple image upscaling techniques are in use, for quite a few years now. Bicubic interpolation (Keys et al., 1981), Super Resolution Convolutional Neural Network (SRCNN) and Rapid and Accurate Image Super Resolution (RAISR) (Romano et al., 2017) are few to mention. SRCNN and RAISR give the best output with the least loss of information density among the others. The performance of SRCNN is better than RAISR but with the cost of greater computation (Dong et al., 2016). Super resolution has significant applications in regular video information enhancement, surveillance, medical diagnosis, astronomical observation, biometric information identification and so on (Yue et al.,2016).

Traditionally any quality enhancement of Satellite images required hardware upgradation of imaging devices in the satellite. This is still the most reliable technique of image quality enhancement. Quality improvement of original multispectral satellite imagery is possible (Starovoitov et al.). They used a technique for multispectral image fusion and minimized the color composite distortion and resolution of the fused images. Deep networks are effective to study low resolution image recognition (Wang et al.,2016). Large scale satellite images can be processed by using convolutional neural network(CNN) (Wang et al., 2016) and accurate pixel wise classification can be generated (Emmanuel et al.,2017). CNN can also be used for per pixel classification and for improving segmentation (Martin et al., 2016).

Michael et al. used Landsat-7 SLC-off image for forest change detection through pixel and polygon based comparison (Michael et al.,2007). They found that using a remediated product such as the histogram- or segment-based gap-filled images increases the ability to accurately detect change. Liebel et al. showed necessary steps to successfully adapt a CNN- based single-image super resolution approach for multispec- tral satellite images (Liebel et al.,2016). Multiple approaches, like K-Means clustering algorithm (Babawuro et al.,2013), object-based approach (Vo et al., 2013), or neural network (Mayank et al.,2005) can be used for land cover classification. Torahi et al. quantified the land cover change patterns in Dehdez area demonstrated the potential of multitemporal Landsat and ASTER data in land management and policy decisions (Torahi et al.,2011). Using the approach of post-classification comparison Akhter et al. found that forest cover changed during the 11 years' period (1989-2000) (Akhter et al.,2006). But the process of classifying the ground features into specific classes introduces thematic errors during the classification process where errors are specifically driven by mixed pixels or spectral confusions. Asner et al. presented an integrated analysis environment to support rapid regional-scale mapping of tropical forests using a variety of common satellite sensors (Asner et al.,2009). They found that management efforts require high-resolution mapping to support policies and forest monitoring efforts. Shopan et al. derived that till 2000 the biomass concentration of the Sundarbans was increasing continuously, but during 2000-2009 the concentration reduced mostly because of massive deforestation using Landsat satellite images (Shopan et al.,2014). Though satellite images have been used for forest monitoring, low resolution of the images\ makes the result error- prone.

Using super resolution to enhance images for better detection of forest land covers, is, so far, a unique approach. And to the best of our knowledge, no study has been conducted so far to explicitly analyze the effect of applying Super Resolution Convolutional Neural Network (SRCNN) for an enhanced recognition of classes within an image. The aim of this paper is to test satellite imagery for pixel by pixel recognition using supervised technique and check if any enhancement of classification can be achieved in upscaled images. We'll test this for two different image upscaling techniques- Bicubic Interpolation and Super Resolution Convolutional Neural Network.

**METHOD AND EQUATIONS:**

**Image Super resolution**

Super resolution imaging is a class of techniques that improves the resolution of an image. It can be described as a nonlinear mapping from a low to high dimensional space. High resolution images are necessary for human interpretation and automatic machine detection. Image resolution describes the details of an image. The higher the resolution, the higher the details. The resolution of an image can be classified in different classes- pixel resolution, spatial resolution, spectral resolution, temporal resolution and radiometric resolution. In our study, we have used two super resolution techniques- bicubic and SRCNN.

**Bicubic Interpolation**

Bicubic interpolation is often chosen over bilinear interpolation when speed is not an issue. Images resembled with bicubic interpolation is less interpolation distortion. Bicubic interpolation considers 16(4x4) pixels where bilinear interpolation takes only 4(2x2) images into account.

Let the function value *f* and the derivatives *fx*, *fy* and *fxy* be known at four corners **(0,0), (1,0), (0,1), (1,1)** of a unit square. The interpolation surface then can be described as-

$$p(x, y) = \sum_{i=0}^{3} \sum_{j=0}^{3} a_{ij} x^i y^j$$

$$f(0,0) = p(0,0) = a_{00}$$

$$f(1,0) = p(1,0) = a_{00} + a_{10} + a_{20} + a_{30}$$

$$f(0,1) = p(0,1) = a_{00} + a_{01} + a_{02} + a_{03}$$

$$f(1,1) = p(1,1) = \sum_{i=0}^{3} \sum_{j=0}^{3} a_{ij}$$

The interpolation problem should determine the 16 coefficients. Four equations are derived by matching *p(x,y)* with the functional values:

**Backpropagation Neural Network**

A scaled conjugate back propagation algorithm (Møller, 1993) is implemented to train the neural network for classifying the land classes from multispectral Satellite images. An artificial neural network is a mimicry of human perception process that takes place with the systematic activation of the actual biological neurons. A basic neural network comprises of one input layer, one output layer and hidden layers in between them. The nodes of different layers are interconnected. Weight of each node are updated every step according to the state of activation of the neuron decided by an activation function. They are recalibrated in such a way that they can reproduce the output layer. Detail understanding of the working principle of artificial neural network has been discussed elsewhere (YEGNANARAYANA, 2009).

**Super Resolution Convolutional Neural Network**

SRCNN produces expanded images with improved details. It creates a bimodal image with improved foreground and background details and allows sharp discontinuity at the edges. SRCNN consists of three convolutional layers. The filter size of first layer is 9x9, second layer is 3x3 and third layer is 1x1. The first layer generates 64 feature maps, the second layer generates 32 and the final layer generates only the output. The low-resolution image is first upscaled to the desired size using bicubic interpolation. Let us denote this interpolation as **Y.** We will generate a high-resolution image **F(Y)** from **Y** by a mapping F which consists of three steps-
1. Patches are extracted from low resolution images and represent them as high dimensional vectors.
2. The high dimensional vectors are non-linearly mapped onto another high dimensional vector.
3. Final high-resolution image is generated by combining the high-resolution patch wise representations.

**Peak Signal-to-noise Ratio**

Image quality assessment is a prior requirement to evaluate the effectiveness of different image processing methods. Peak Signal to Noise Ratio (PSNR) is a commonly used tool to quantify a signal quality with reference to the original signal and have been used in many instances including satellite images (Demirel and Anbarjafari, 2010), CT and MR images (Ali et al., 2008) and others. A handful number of image quality metrics have been developed and tested so far, but reportedly they show no clear advantage over mathematically simple metrics like PSNR (Wang et al., 2002).

For a MxN reference image f and a test image g of the same size and class, PSNR is calculated as –

$$PSNR(f,g) = 10\log_{10}(\frac{255^2}{MSE(f,g)})$$

Where MSE (f, g) (Mean Square Error) is defined as –

$$MSE(f,g) = \frac{1}{MN}\sum_{i=1}^{M}\sum_{j=1}^{N}(f_{ij}-g_{ij})^2$$

PSNR corresponds to the mean square error between reference and test image. This MSE is often regarded as the noise induced in the test image due to processing. Thus, a higher value of PSNR means there is limited deviation of quality from the reference image to test image when restored. When the test image and the reference image are identical, MSE is zero, and PSNR becomes infinite

**Study sample**

One set of LANDSAT 7 ETM+ data are acquired from USGS Earth Explorer achieve, for 6 different spectral bands. The study site is chosen to be the eastern Sundarbans (centered at 21° 40'N 90° 2'E). Band 1 to 5 and 7 are used. They are blue, green, red, near infrared, shortwave infrared-1 and 2 respectively. The sample dataset is chosen on the basis of least cloud coverage.

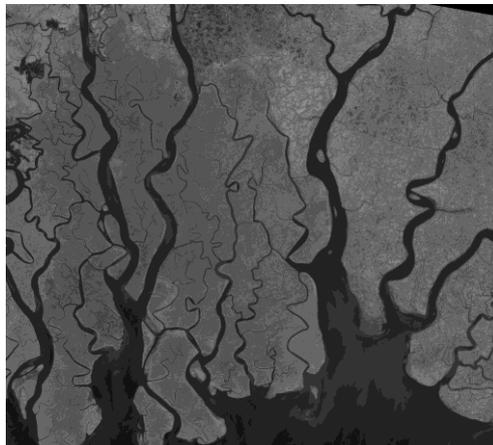

Figure 0: Band-4 of the six bands used in the study, WRS path 137 row45 (a) LANDSAT 7 ETM+ image Band -4 (Near Infrared), 7[th] November, 1999.

**Methodology**

Areas, that are well representative of three study classes (deep forest, light forest and river), are chosen from within the study site. All the images are then segmented into 3x3 size patches and then to columns. Each column can act as an input neuron to the classification algorithm. Patch pixels from all six bands are ordered and fed to a feed forward back propagation neural network as the training dataset. About 25000 neurons were fed feom each classes where each neuron has 24 nodes. Hereafter, Super Resolution Convolution network and bicubic interpolation methods are implemented to upscale the test images. Then the previously trained network was assigned to detect classes in the test images. For the demonstration purpose, we segmented the image into three classes- Deep Forest (ash color), Light Forest (white), River(black).

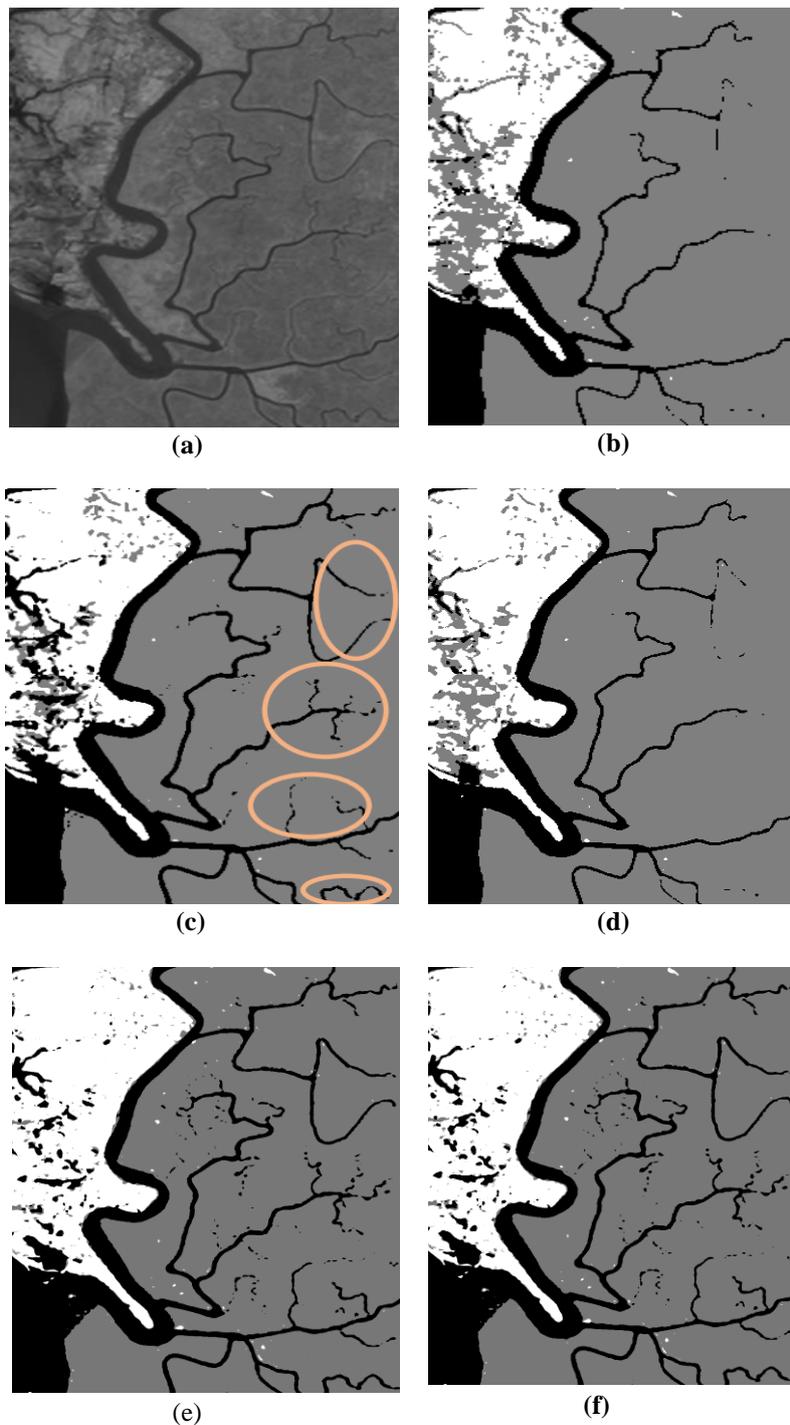

**Figure 1:** (a) Original image. Blue, Red, Green, NIR, SWIR-1 and 2 bands are used in the study. NIR band is presented here. (b) detection of classes from original image (c) detection of classes from 3 times upscaled image (SRCNN) (d)) detection of classes from 3 times upscaled image (bicubic interpolation) (e) detection of classes from

27 times upscaled image (SRCNN) (f) detection of classes from 27 times upscaled image (bicubic interpolation)

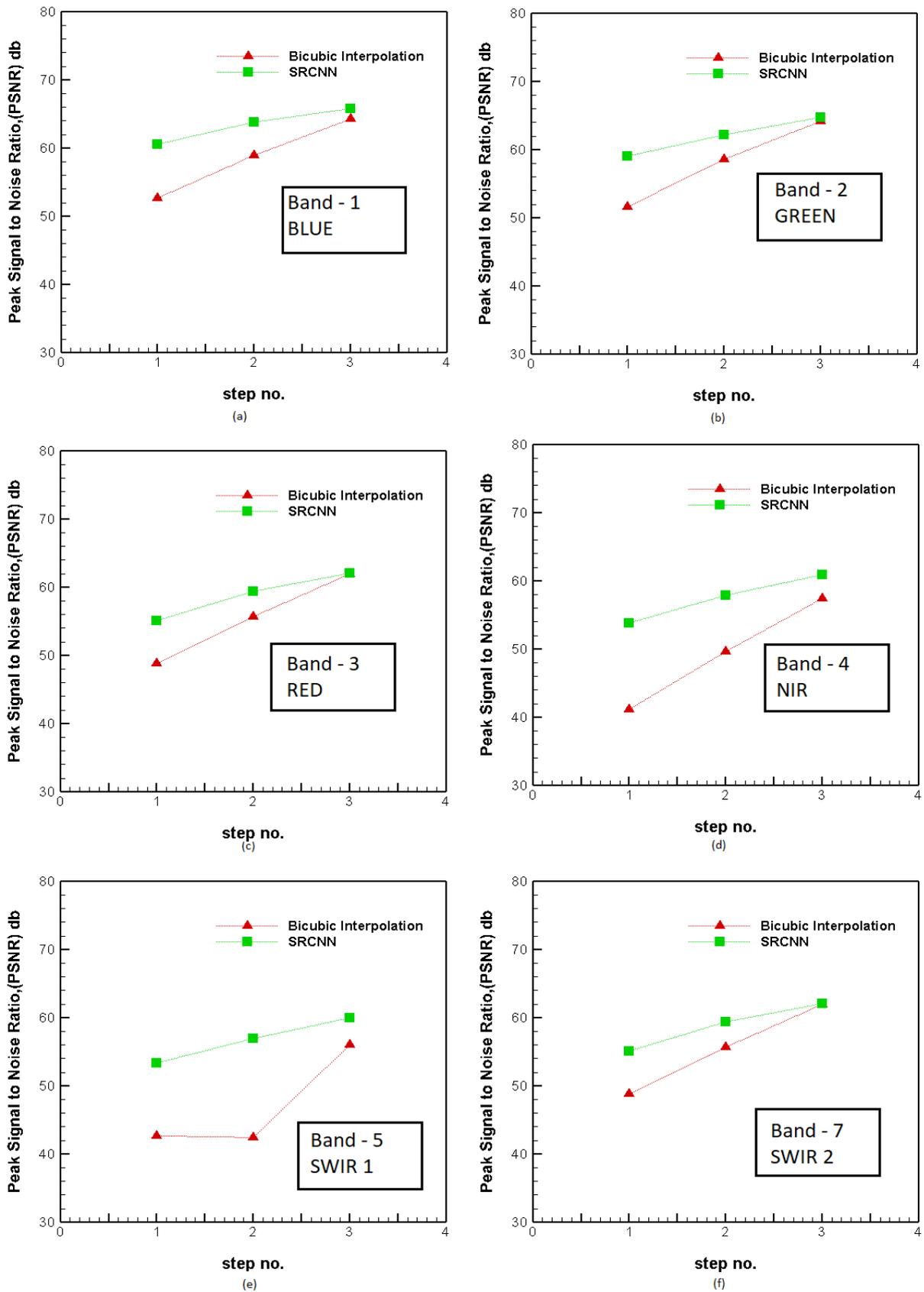

Fig 2. Variation of Peak Signal to Noise Ratio for LANDSAT 7 ETM+ Bands 1-6 and 7 with the change of degree of image upscaling for SRCNN and Bicubic interpolation method. Any nth step corresponds to a 3 times up-sampling

of image and each PSNR corresponds to the PSNR with reference to the (n-1)th step image.
**RESULT AND DISCUSSION:**

Does an up-scaled image perform better in regards of classification problem? To address this question we arranged an experiment in the following way- A classifier is trained with a fixed set of training data and then it is assigned the task of classifying features in a set of original satellite images along with the images that are scaled up to different degree in the shade of Bicubic Interpolation and SRCNN techniques. It is to be mentioned that, before feeding to the network input layer, each image is pooled by 2x2 pixel$^2$ areas to ensure the local features for each class to be trained. Exactly in the same way, in the case of testing the image for feature detection, 2x2 pooling is performed in the test images as well. The pooled area is replaced by the classifier with pixels of s single value. This is how information is lost in the flow. And when the features in the test images are one or two pixels across, they may be pooled with nearby large features. Thus, when the pooled pixels are fed to the network, they are on a high vulnerability of getting misclassified. When an image is scaled up by some technique, any single pixel is replaced by a population of relevant pixels. Now the quality of up-scaling technique comes to play their crucial role. In case of traditional super-resolution techniques, like, Bilinear and bicubic interpolations, any pixel is populated by a calculated gradient. This is why the output becomes blurry and the blurry gradient of pixels can only represent a feature by chance and they go incorrectly classified. To be precise, these techniques have no inherent tendency to preserve a feature in the image which is the main shortcoming as their use for enhancement of feature classification. Again, if sharpening task is done on the images, they get contaminated with unnecessary noise which again hampers the quality of feature detection. However, super resolution convolution network is a deep network, developed by dong et al., learns the mapping between low and high-resolution images very efficiently and their performance is excellent in terms of preserving the features in the upscaled images. Thus, we hypothesized that images upscaled by this technique will perform better in extracting the tiniest features that go misclassified otherwise. Figure 1 is a representation of one of our test results. Figure 1(a) is the image data of one of the six LANDSAT-7 bands (band 4) we used in our study. Figure 1(b) is the output of our classifier that derived it from the original set of images. Figure 1(d) was derived by the classifier, but now on a three times up-scaled image. Upscaling in the test image of figure 1(b) was performed in Bicubic interpolation technique. Classes of figure 1(b) was detected from a 3 times up-scaled image again, but now, done by the SRCNN technique. It is clearly seen that the feature extraction in figure 1(b) and 1(d) fails to correctly classify the tiny features, like, very narrow rivers branches. However, they were far more correctly classified when the upscaling is performed in SRCNN (fig1(c)). Now figure 1(e) and 1(f) demonstrated the most interesting feature of our study. It shows at about27th times upscaling, very similar and detailed features could be extracted from both bicubic and SRCNN images.

The bands used in the study are tested for Peak signal-to-noise ratio to assure the restoration quality. At first the test sample area (375x516 pixels) from original images are collected from USGS website are upscaled to 645x888 pixel using two fundamentally different image upscaling approaches- Bicubic Interpolation and SRCNN. Now Peak signal to noise ratio is calculated for each band of the two groups of up-scaled image set. In this study, for any n$^{th}$ order up-scaled image the PSNR is calculated with respect to the image derived by (n-1)$^{th}$ order of SRCNN upscaled image. We derived the bicubic and SRCNN image for 3, 9 and 27 times of the original image and calculated the PSNR for each of them. The calculated PSNR are presented in figure 3. Figure 3 shows an interesting trend for PSNR. It seems obvious that both for Bicubic interpolation and Super Resolution Convolutional Neural network, PSNR rises with the derived image size. For any step presented in figure 3, the derived image size is $3^{step}$ of the original image. Thus, PSNR maintains a power relationship with image size. This relationship holds true for all the bands observed. It is further observed that, PSNR is observed higher in case of SRCNN processed images than their corresponding bicubic ones. This observation is important, and holds a key for the validity of hypothesis we are proving in this paper. Figure 3 shows again that the quality gap between SRCNN and bicubic interpolation decreases with up-scaled image size. And for about 27$^{th}$ times increase of original image, results from both technique almost agrees with each other. These reminds us of a saturation of image quality at these points.

**CONCLUSION:**

From our study, we can draw several interesting conclusive remarks-
Firstly, we hypothesized that performance of image classification algorithms would be better if they were applied to an upscaled image instead of their original versions. From our study, it was discovered that, the quality of image upscaling algorithms is equally important in that case. And scaled up images that are derived from high performance super resolution algorithms like SRCNN, in fact, validated our hypothesis. Satellite images, that were scaled up by SRCNN, showed very promising result. After using SRCNN, very tiny features in the satellite image were correctly classified which were misclassified under general conditions. Hence in our concerned problem, SRCNN outperformed Bicubic interpolation method very clearly. It was also discovered that, most features were revealed at a moderate level of image upscaling by SRCNN. Very high order of upscaling may not be required for practical use. For very high upscaling, SRCNN and Bicubic interpolation performs in the similar order. This is a major finding of our study. Again, in this study we observed the variation of peak signal-to-noise ratio with the variation of up-scaled image. It was found that images processed by SRCNN holds better PSNR, and the increase of PSNR with image size has a power relationship. However, at very higher image size the PSNR gets saturated and coincides for both methods.

**ACKNOWLEDGEMENT** Our sincere thanks to Satyajit Mojumder sir, lecturer, Bangladesh University of Engineering and Technology, for supervising the course ME-262, upon which a significant portion of the coding for this research is based. This manuscript was originally prepared for 38th Asian Conference on Remote Sensing, 2017.

**REFERENCES:**

A. A. Shopan, G. T. Islam, A. S. Islam, M. M. Rahman, A. N. Lazár, C. Hutton, Temporal variation of biomass concentration in the bangladesh sundarbans using remote sensing techniques, in: Internat Conf on Adv in, Vol. 105, 2014, p. l.

A. A. Torahi, S. C. Rai, et al., Land cover classification and forest change analysis, using satellite imagery-a case study in dehdez area of zagros mountain in iran, Journal of Geographic Information System 3 (01) (2011) 1. doi:10.4236/jgis.2011.31001.

B. Usman, Satellite imagery land cover classification using k-means clustering algorithm computer vision for environmental information extraction, Elixir Computer Science and Engineering 63 (2013) 18671–5.

B. Yegnanarayana, Artificial neural networks, PHI Learning Pvt. Ltd., 2009.

C. Dong, C. C. Loy, K. He, X. Tang, Learning a deep convolutional network for image super-resolution, in: European Conference on Computer Vision, Springer, 2014, pp. 184–199. doi:10.1007/978-3-319-10593-213.

C. Dong, C. C. Loy, K. He, X. Tang, Image super-resolution using deep convolutional networks, IEEE transactions on pattern analysis and machine intelligence 38 (2) (2016) 295–307. doi:10.1109/TPAMI.2015.2439281.

D. L. Williams, S. Goward, T. Arvidson, Landsat, Photogrammetric Engineering & Remote Sensing 72 (10) (2006) 1171–1178. doi:10.14358/PERS.72.10.1171.

E. Maggiori, Y. Tarabalka, G. Charpiat, P. Alliez, Convolutional neural networks for large-scale remote-sensing image classification, IEEE Transactions on Geoscience and Remote Sensing 55 (2) (2017) 645–657. doi:10.1109/TGRS.2016.2612821.

G. P. Asner, D. E. Knapp, A. Balaji, G. Paez-Acosta, et al., Automated mapping of tropical deforestation and forest degradation: Claslite, Journal of Applied Remote Sensing 3 (1) (2009) 033543. doi:10.1117/1.3223675.

H. Demirel, G. Anbarjafari, Satellite image resolution enhancement using complex wavelet transform, IEEE geoscience and remote sensing letters 7 (1) (2010) 123–126. doi:10.1109/LGRS.2009.2028440.

L. Liebel, M. Korner, Single-image super resolution for multispectral remote sensing data using convolutional neural networks, in: XXIII ISPRS Congress proceedings p883-890, 2016. doi:10.5194/isprs-archives-XLI-B3-883-2016.

L. Yue, H. Shen, J. Li, Q. Yuan, H. Zhang, L. Zhang, Image super-resolution: The techniques, applications, and future, Signal Processing 128 (2016) 389–408. doi:10.1016/j.sigpro.2016.05.002.

M. Akhter, Remote sensing for developing an operational monitoring scheme for the sundarban reserved forest, bangladesh, Ph.D. thesis, Department of World Forestry, University of Hamburg (2006).

M. A. Wulder, S. M. Ortlepp, J. C. White, S. Maxwell, Evaluation of landsat-7 slc-off image products for forest change detection, Canadian Journal of Remote Sensing 34 (2) (2008) 93–99. doi:10.5589/m08-020.

M. F. Møller, A scaled conjugate gradient algorithm for fast supervised learning, Neural networks 6 (4) (1993) 525–533. doi:10.1016/S0893-6080(05)80056-5.

M. Langkvist, A. Kiselev, M. Alirezaie, A. Loutfi, Classification and segmentation of satellite orthoimagery using convolutional neural networks, Remote Sensing 8 (4) (2016) 329. doi:10.3390/rs8040329.

M. Toshniwal, Satellite image classification using neural networks, in: 3rd International Conference: Sciences of Electronic, Technologies of Information and Telecommunications, 2005, pp. 27–31. doi:10.1109/ICNSC.2005.1461193.

Q. T. Vo, N. Oppelt, P. Leinenkugel, C. Kuenzer, Remote sensing in mapping mangrove ecosystemsan object-based approach, Remote Sensing 5 (1) (2013) 183–201. doi:10.3390/rs5010183.

R. Keys, Cubic convolution interpolation for digital image processing, IEEE transactions on acoustics, speech, and signal processing 29 (6) (1981) 1153–1160. doi:10.1109/TASSP.1981.1163711.

V. Starovoitov, A. Makarau, Multispectral image pre-processing for interactive satellite image classification.

Z. Wang, A. C. Bovik, L. Lu, Why is image quality assessment so difficult?, in: Acoustics, Speech, and Signal Processing (ICASSP), 2002 IEEE International Conference on, Vol. 4, IEEE, 2002, pp. IV–3313.

Z. Wang, S. Chang, Y. Yang, D. Liu, T. S. Huang, Studying very low resolution recognition using deep networks, in: Proceedings of the IEEE Conference on Computer Vision and Pattern Recognition, 2016, pp.4792–4800.